\documentclass[conference]{IEEEtran}
\IEEEoverridecommandlockouts
\usepackage{cite}
\usepackage{amsmath,amssymb,amsfonts}
\usepackage{float}
\usepackage{eso-pic}
\usepackage{enumitem}
\usepackage[ruled,vlined]{algorithm2e}
\usepackage{graphicx}
\usepackage{tikz}
\usetikzlibrary{fit, positioning, arrows.meta}
\usepackage{booktabs}
\usepackage{multirow}
\usepackage{textcomp}
\usepackage{xcolor}
\usepackage{url}
\usepackage[colorlinks=true, linkcolor=black, citecolor=black, urlcolor=black]{hyperref}

\def\BibTeX{{\rm B\kern-.05em{\sc i\kern-.025em b}\kern-.08em
    T\kern-.1667em\lower.7ex\hbox{E}\kern-.125emX}}

\begin{document}

\title{Multimodal Foundation Models for Early Disease Detection\\

}

\author{\IEEEauthorblockN{\textsuperscript{} Md Talha Mohsin}
\IEEEauthorblockA{
\textit{The University of Tulsa} \\
800 S Tucker Dr, Tulsa, OK 74104, USA. \\
}
\and
\IEEEauthorblockN{\textsuperscript{} Ismail Abdulrashid}
\IEEEauthorblockA{
\textit{The University of Tulsa} \\
800 S Tucker Dr, Tulsa, OK 74104, USA. \\
}
}

\maketitle

\begin{abstract}

Healthcare data now span EHRs, medical imaging, genomics, and wearable sensors, but most diagnostic models still process these modalities in isolation. This limits their ability to capture early, cross-modal disease signatures. This paper introduces a multimodal foundation model built on a transformer architecture that integrates heterogeneous clinical data through modality-specific encoders and cross-modal attention. Each modality is mapped into a shared latent space and fused using multi-head attention with residual normalization. We implement the framework using a multimodal dataset that simulates early-stage disease patterns across EHR sequences, imaging patches, genomic profiles, and wearable signals, including missing-modality scenarios and label noise. The model is trained using supervised classification together with self-supervised reconstruction and contrastive alignment to improve robustness. Experimental evaluation demonstrates strong performance in early-detection settings, with stable classification metrics, reliable uncertainty estimates, and interpretable attention patterns. The approach moves toward a flexible, pretrain-and-fine-tune foundation model that supports precision diagnostics, handles incomplete inputs, and improves early disease detection across oncology, cardiology, and neurology applications.
\end{abstract}


\section{Introduction}

The increasing availability of heterogeneous patient data creates both analytical challenges and significant opportunities for data-driven healthcare. Modern clinical data include diverse modalities such as medical imaging, longitudinal EHR records and wearable sensor streams, auditory and video data, unstructured clinical text, and molecular measurements including genomics and proteomics \cite{jmir_multimodal_2024}. EHR systems further provide detailed longitudinal patient histories across encounters, capturing diagnoses, medications, and procedures that support temporal modeling of clinical trajectories \cite{choi_retain_2016, ma_dipole_2017}. As precision medicine becomes increasingly central to clinical practice {\cite{ma_kame_2018}}, effective clinical decision-making depends on synthesizing these heterogeneous information sources. Yet most predictive models rely on a single modality, limiting their ability to capture cross-modal dependencies and reducing their potential for patient-centric insights.

Recent advances in deep learning architectures have facilitated the learning of complex, non-linear relationships across heterogeneous data modalities.For instance, attention-based architectures can dynamically weight the most relevant signals and, when pretrained on large heterogeneous datasets, learn generalizable representations for downstream clinical applications {\cite{sun_medical_2025}}. This foundation-model paradigm, established in NLP and computer vision {\cite{bommasani_foundation_2021}}, motivates multimodal approaches to early disease detection. At the same time, transformer architectures have shown very strong performance in large-scale pretraining, learning contextual representations transferable to many downstream tasks {\cite{yang_xlnet_2019}}. Foundation models extend such benefits by offering broad generalization and prompting capabilities {\cite{azad_foundational_2023}}. In light of these trends, we suggest a transformer-based multimodal foundation model that combines data from electronic health records (EHR), imaging, genomics, and wearable sensors. In this framework, cross-modal attention lets the model see how different modalities are related to each other and then lets it learn from a wide range of tasks, resulting in representations that can be used for many different patients. This unified framework seeks to enhance integrated diagnostic tools that facilitate precision medicine by amalgamating diverse clinical data streams within a singular computational architecture.

\section{Related Work}

\subsection{Multimodal Integration}

The wide use of EHR systems has led to the development of predictive models that can support clinical care {\cite{choi_retain_2016}. As healthcare data now extend far beyond EHRs to include imaging, genomics, and signals from wearable devices, these models are increasingly applied to improve patient outcomes {\cite{choi_gram_2017}. However, most current methods remain limited to a single data type; so they miss important cross-modal information, struggle with incomplete inputs, and often rely on specialized fusion methods that do not scale well to high-dimensional biomedical data. In practice, clinical information is multimodal. Integrating different sources of data is often necessary for accurate diagnosis and effective treatment {\cite{jmir_multimodal_2024}. Studies show that combining EHRs with imaging or genomics can improve diagnostic accuracy when paired with deep learning methods {\cite{ehr_knowgen_2024}. At the same time, foundation models—large pretrained systems designed to adapt across many tasks {\cite{bommasani_foundation_2021} —are becoming increasingly important in healthcare for their ability to generalize across data types and domains  {\cite{azad_foundational_2023},{\cite{ieee_foundation_healthcare_2024}. Together, these developments suggest a natural convergence: Multimodal foundation models provide a way to improve early disease detection and enable more precise, patient-centered care.

\subsection{Attention-Based Models}

Attention-based transformer models came out in 2017 {\cite{vaswani_attention_2017} and quickly gained popularity for working with sequential data. They handle sequences differently from older models like Recurrent Neural Networks (RNNs), which process data step by step. Instead, transformers can consider the whole sequence at once, which tends to make it easier to spot patterns that span long sections of the data {\cite{khan_transformers_2022}. In addition, they scale fairly well when the dataset is large and can be trained in parallel. That combination of flexibility and efficiency is one reason researchers have started using them in many areas, including healthcare {\cite{wolf_transformers_2020}.

In healthcare, transformers have been applied to handle large and messy data sources, especially clinical notes and records. They can highlight which pieces of information matter most for a given prediction, making them useful for tasks where not every variable has the same importance {\cite{denecke_transformer_2024}. This property is also what makes them appealing for multimodal learning: the model can weigh structured EHR entries against imaging, genomic profiles, or even wearable sensor data without relying on heavy feature engineering \cite{jmir_task_specific_2024}. 

We address the gaps of single-modality modeling, missing cross-modal fusion, and absent foundation-style pretraining by building transferable patient representations that capture subtle early disease patterns.

\section{Proposed Framework}

We propose a multimodal transformer-based framework to combine diverse biomedical data sources for early illness diagnosis. Unlike unimodal techniques, which handle only one type of input, our architecture combines a variety of clinical data streams, such as EHRs, medical imaging, genomic sequences, and wearable sensor data. The architecture stresses both flexibility and robustness, allowing predictions even when some modalities are absent, as well as extensibility to new patient data sources that were not available during training. Figure \ref{fig:multimodal_arch} shows a patient-centric multimodal transformer design.

\tikzstyle{layer} = [rectangle, draw, rounded corners, thick, minimum width=3.5cm, minimum height=1.2cm, align=center]
\tikzstyle{subblock} = [rectangle, draw, rounded corners, thin, minimum width=0.9cm, minimum height=0.3cm, font=\tiny, align=center]

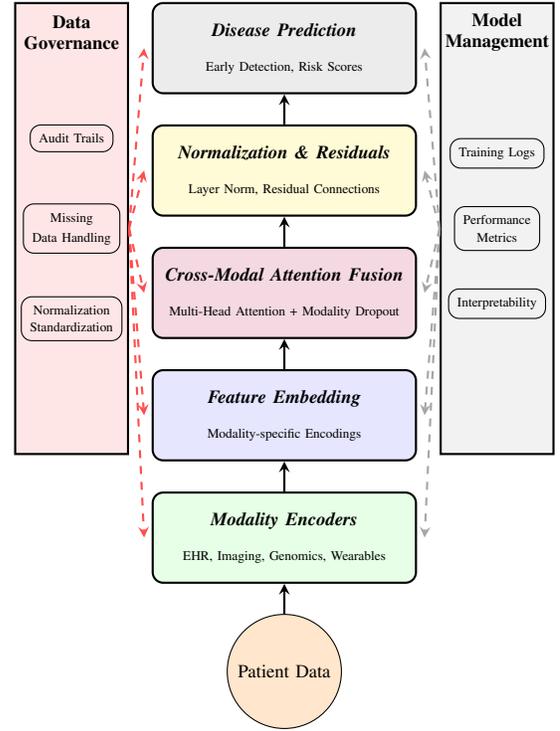
\begin{figure}[h]
\centering
\begin{tikzpicture}[node distance=0.4cm, >=stealth]

\node[draw, circle, fill=orange!20, minimum size=1cm] (patient) {\scriptsize Patient Data};

\node[layer, fill=green!10, above=of patient] (encoders) {\textbf{\textit{\scriptsize Modality Encoders}}\\[1pt]\tiny EHR, Imaging, Genomics, Wearables};

\node[layer, fill=blue!10, above=of encoders] (embedding) {\textbf{\textit{\scriptsize Feature Embedding}}\\[1pt]\tiny Modality-specific Encodings};
\node[layer, fill=purple!15, above=of embedding] (fusion) {\textbf{\textit{\scriptsize Cross-Modal Attention Fusion}}\\[1pt]\tiny Multi-Head Attention + Modality Dropout};
\node[layer, fill=yellow!20, above=of fusion] (norm) {\textbf{\textit{\scriptsize Normalization \& Residuals}}\\[1pt]\tiny Layer Norm, Residual Connections};
\node[layer, fill=gray!15, above=of norm] (prediction) {\textbf{\textit{\scriptsize Disease Prediction}}\\[1pt]\tiny Early Detection, Risk Scores};

\foreach \source/\target in {patient/encoders, encoders/embedding, embedding/fusion, fusion/norm, norm/prediction}
    \draw[->, thick] (\source.north) -- (\target.south);

\node[rectangle, draw, thick, fill=red!10, minimum width=1.5cm, minimum height=6cm,
      anchor=north east] (governance) at ([xshift=-0.3cm]prediction.north west) {};
\node at (governance.north) [yshift=-0.4cm, font=\scriptsize\bfseries, align=center] {\shortstack{Data\\Governance}};
\node[subblock] at (governance.center) [yshift=1.2cm] {Audit Trails};
\node[subblock] at (governance.center) [yshift=0cm] {\shortstack{Missing\\Data Handling}};
\node[subblock] at (governance.center) [yshift=-1.2cm] {\shortstack{\tiny Normalization\\\tiny Standardization}};

\foreach \layer in {prediction, norm, fusion, embedding, encoders}
    \draw[->, thick, dashed, red!70] (governance.east) -- ([xshift=-0.1cm]\layer.west);

\node[rectangle, draw, thick, fill=gray!10, minimum width=1.5cm, minimum height=6cm,
      anchor=north west] (monitoring) at ([xshift=0.3cm]prediction.north east) {};
\node at (monitoring.north) [yshift=-0.4cm, font=\scriptsize\bfseries, align=center] {\shortstack{Model\\Management}};
\node[subblock] at (monitoring.center) [yshift=1cm] {Training Logs};
\node[subblock] at (monitoring.center) [yshift=0cm] {\shortstack{Performance\\Metrics}};
\node[subblock] at (monitoring.center) [yshift=-1cm] {Interpretability};

\foreach \layer in {prediction, norm, fusion, embedding, encoders}
    \draw[->, thick, dashed, gray!70] (monitoring.west) -- ([xshift=0.1cm]\layer.east);

\end{tikzpicture}
\caption{Multimodal transformer architecture}
\label{fig:multimodal_arch}
\end{figure}

\subsection{Problem Definition}

Let each patient record be denoted by:
\[
\mathcal{P}_i = \{X^{ehr}_i, X^{img}_i, X^{gen}_i, X^{sens}_i\},
\]
where $X^{ehr}_i \in \mathbb{R}^{T \times d_{ehr}}$ represents temporal EHR data, 
$X^{img}_i \in \mathbb{R}^{H \times W \times C}$ corresponds to imaging modalities, 
$X^{gen}_i \in \mathbb{R}^{L_{gen} \times d_{gen}}$ denotes genomic features, and 
$X^{sens}_i \in \mathbb{R}^{T_s \times d_{sens}}$ encodes wearable sensor signals.  

The predictive objective is:
\[
\hat{y}_i = f_\theta(\mathcal{P}_i), \quad y_i \in \{0,1\}^K,
\]
where $\hat{y}_i$ denotes the probability distribution over $K$ disease classes.  

As not all patients will have complete data across all modalities, our framework is designed to accommodate incomplete records by incorporating modality dropout during training as well as enabling inference on any available subset of $\{X^{ehr}, X^{img}, X^{gen}, X^{sens}\}$> It also ensures practical applicability in clinical settings where data coverage is uneven.  

\subsection{Modality-Specific Encoders}

Each of the modality is transformed into a latent representation by a dedicated encoder:
\[
h^{m}_i = \phi_m(X^{m}_i), \quad m \in \{ehr, img, gen, sens\}.
\]

\begin{algorithm}[h]
\caption{Modality Encoding}
\KwIn{Patient data $\mathcal{P} = \{X^{ehr}, X^{img}, X^{gen}, X^{sens}\}$}
\KwOut{Latent representations $\{h^{ehr}, h^{img}, h^{gen}, h^{sens}\}$}
\ForEach{modality $m \in \{ehr, img, gen, sens\}$}{
    $h^m \leftarrow \phi_m(X^m)$ \tcp*{Apply modality-specific encoder}
}
\Return{$\{h^{ehr}, h^{img}, h^{gen}, h^{sens}\}$}
\end{algorithm}

Here, raw inputs are grouped together using feature spaces. Sequential embeddings are used to capture temporal dependencies for electronic health records (EHR); Convolutional Neural Networks (CNNs) or Vision Transformers (ViTs) are used to extract spatial hierarchies from imaging data; sequence models such as 1D CNNs are used to encode the data for genomics; and temporal CNNs or Gated Recurrent Units (GRUs) are used to model the data for wearable signals. Each encoder learns modality-specific representations $h^m$ that ensure that all input types are identical while preserving significant features.  

\subsection{Cross-Modal Attention Fusion}

Encoded features are aggregated to form a unified embedding. Let:
\[
Z = \{h^{ehr}_i, h^{img}_i, h^{gen}_i, h^{sens}_i\}.
\]

Cross-modal attention operates over this set:
\[
\text{Attention}(Q,K,V) = \text{softmax}\!\left(\frac{QK^\top}{\sqrt{d_k}}\right) V,
\]
with
\[
Q = W_Q Z, \quad K = W_K Z, \quad V = W_V Z.
\]

The fused embedding is:
\[
h^{fusion}_i = \text{Concat}(\text{head}_1, \dots, \text{head}_H) W_O.
\]

\begin{algorithm}[h]
\caption{Cross-Modal Fusion}
\KwIn{Encoded representations $Z = \{h^{ehr}, h^{img}, h^{gen}, h^{sens}\}$}
\KwOut{Fused embedding $h^{fusion}$}
$Q \leftarrow W_Q Z, \; K \leftarrow W_K Z, \; V \leftarrow W_V Z$\;
\For{head $= 1$ to $H$}{
    $head_h \leftarrow \text{softmax}\!\left(\frac{QK^\top}{\sqrt{d_k}}\right) V$\;
}
$h^{fusion} \leftarrow \text{Concat}(head_1, \dots, head_H) W_O$\;
\Return{$h^{fusion}$}
\end{algorithm}

Here representations across modalities are integrated and queries, keys, and values are projected from the modality embeddings. Multi-head attention ensures that the model can capture distinct relationships (e.g., correlating imaging abnormalities with lab results or linking genetic variants to wearable data patterns). The concatenation of attention heads followed by a linear projection yields $h^{fusion}$, a joint patient embedding suitable for downstream classification.  

\subsection{Training Strategy}

The training protocol consists of two stages: first, a large corpus of multimodal data is used for self-supervised pretraining and second, the pretrained model is then fine-tuned with supervised labels for disease prediction.  Objectives of the training phase include: (i) masked reconstruction of missing inputs and (ii) cross-modal contrastive learning to align paired modalities.

\begin{algorithm}[h]
\caption{Training Procedure}
\KwIn{Unlabeled data $\mathcal{D}_{pre}$, labeled data $\mathcal{D}_{task}$}
\KwOut{Optimized parameters $\theta^*$}
\ForEach{batch $\mathcal{B} \subset \mathcal{D}_{pre}$}{
    Compute reconstruction loss $\mathcal{L}_{mask}$\;
    Compute contrastive loss $\mathcal{L}_{contrast}$\;
    Update $\theta$ using $\nabla(\mathcal{L}_{mask} + \alpha \mathcal{L}_{contrast})$\;
}
\ForEach{batch $\mathcal{B} \subset \mathcal{D}_{task}$}{
    Encode and fuse modalities\;
    Predict disease label $\hat{y}$\;
    Compute supervised loss $\mathcal{L}_{task} = \text{CE}(y, \hat{y})$\;
    Update $\theta$ with gradient descent\;
}
\Return{$\theta^*$}
\end{algorithm}

The encoders and fusion module learn general multimodal patterns during pretraining. This helps the model deal with data that is noisy. On the other hand, masked reconstruction pushes it to fill in missing information, while contrastive learning makes embeddings from different modalities fit together. During the fine-tuning stage, the network is made to work better at finding diseases by optimizing cross-entropy loss on labeled data. This two-step process finds a middle ground between having a lot of general knowledge and doing well on specific tasks.

\subsection{End-to-End Reasoning Flow}

\renewcommand{\thealgocf}{}
\begin{algorithm}[h]
\caption{Chain-of-Thought Template: Multimodal Diagnostic Inference}
\label{alg:Chain-of-Thought}
\KwIn{Patient record $\mathcal{P}=\{X^{ehr},X^{img},X^{gen},X^{sens}\}$ (some modalities may be missing)}
\KwOut{Prediction $\hat{y}$, uncertainty $u$, explanation $E$}

\SetKwFunction{Encode}{Encode}
\SetKwFunction{Embed}{Embed}
\SetKwFunction{Fuse}{CrossModalFuse}
\SetKwFunction{Predict}{PredictHead}
\SetKwFunction{Uncert}{EstimateUncertainty}
\SetKwFunction{Update}{ApplyFeedback}

\tcp{Input acquisition \& preprocessing}
$X^{m} \leftarrow \text{Preprocess}(X^{m})\;\forall m \in \{ehr,img,gen,sens\}$\;
Simulate/record missing-modality mask $M$\;

\tcp{Unimodal representation learning}
\ForEach{modality $m$ available}{
    $z^{m} \leftarrow$ \Encode{$X^{m}$} \tcp*{encoder: Transformer / CNN / 1D-CNN / TCN}
    $e^{m} \leftarrow$ \Embed{$z^{m}$} \tcp*{project to shared latent space}
}

\tcp{Cross-modal alignment (optional contrastive step during pretraining)}
Align $\{e^{m}\}$ with contrastive or projection losses (pretraining)\;

\tcp{Multimodal reasoning via fusion}
$h^{(0)} \leftarrow \text{Aggregate}(\{e^{m}\})$ \tcp*{e.g., concatenation or learned pooling}
\For{$\ell = 1$ \KwTo $L$}{
    $h^{(\ell)} \leftarrow$ \Fuse$(h^{(\ell-1)}, \{e^{m}\}, M)$ \tcp*{multi-head cross-modal attention + residuals}
    Optional: apply LayerNorm and Feed-Forward block
}

\tcp{Prediction \& uncertainty}
$\hat{y} \leftarrow$ \Predict$(h^{(L)})$\;
$u \leftarrow$ \Uncert$(h^{(L)}, \hat{y})$ \tcp*{e.g., MC-dropout, ensemble, or Bayesian head}

\tcp{Explanation / Chain-of-Thought trace}
$E \leftarrow \text{ExtractAttentionMaps}( \{ \text{attention weights from } \ell\})$\;

\tcp{Feedback / continual update (deployed system)}
\uIf{feedback available (label/outcome)}{
    \Update$(\theta, \text{feedback})$ \tcp*{fine-tune or federated update}
}

\Return{$\hat{y}, u, E$}
\end{algorithm}

This the chain-of-thought template that depicts the proposed framework's end-to-end reasoning process to supplement the modality-specific algorithms. This template shows how the whole thing works as a diagnostic workflow. The process starts with raw data preprocessing and modality-specific encoding, then moving on to cross-modal fusion and prediction, uncertainty estimates, explanation creation, and updates driven by feedback. This approach aligns with prior provenance-aware digital-twin systems that integrate explainability and lifecycle traceability for clinical AI  \cite{mohsin_paxdt_2025}.

\section{Multimodal Model Design and Learning Framework}

Our proposed model combines different types of clinical data into a single latent space and uses both supervised and self-supervised goals to make it more robust when the data is noisy or missing.

\subsection{Multimodal Data}

We simulate a multimodal dataset comprising four heterogeneous modalities: (1) temporal EHR sequences, (2) $32\times 32$ imaging patches, (3) 500-dimensional genomic profiles, and (4) 3-channel wearable sensor time-series. Each of the modality we used is generated with structured stochasticity to approximate early-stage disease signatures. Positive cases contain weak, partially overlapping signals (e.g., soft EHR trends, mild genomic up-regulation, or subtle imaging lesions), while negative cases occasionally exhibit pseudo-pathological artifacts to induce class overlap.

At the same time, to mimic real-world uncertainty, we introduce $10\%$ label noise and apply a $30\%$ random missing-modality dropout, which produces patient records containing arbitrary subsets of modalities. Table~\ref{tab:params} summarizes key data-generation parameters.

\subsection{Model Architecture}

Each modality is encoded via a dedicated neural encoder mapping raw inputs to a shared 64-dimensional latent space: a GRU for EHR, a two-stage CNN for imaging, a two-layer MLP for genomics, and a temporal convolutional network for wearable signals. Formally,
\[
h_{m} = f_{m}(X_{m}),\quad h_{m} \in \mathbb{R}^{64},\quad m\in\{\text{ehr,img,gen,wear}\}.
\]

The embeddings are stacked and passed through a multihead self-attention transformer that performs cross-modal message passing:
\[
Z = \text{Transformer}\left([h_{\text{ehr}},h_{\text{img}},h_{\text{gen}},h_{\text{wear}}]\right),
\]
followed by mean pooling to obtain the fused representation $z\in\mathbb{R}^{64}$. Attention weights are retained for interpretability.

Self-supervised components include:  
(1) modality masking, where one or more $X_m$ are zeroed-out during training;  
(2) reconstruction decoders $g_m(z)$ trained via $\ell_2$ losses; and  
(3) a CLIP-style contrastive alignment objective applied to cross-modal pairs:
\[
\mathcal{L}_{\text{con}}=
\frac{1}{2}\left[
\text{CE}(S, I)+\text{CE}(S^\top, I)
\right],
\quad 
S_{ij}= \frac{\tilde h_{\text{ehr},i}^\top \tilde h_{\text{img},j}}{\tau}.
\]

A two-layer MLP with Monte Carlo dropout produces final predictions and provides uncertainty estimates via stochastic forward passes.

\subsection{Training Procedure}

Training together lowers the supervised cross-entropy loss, the reconstruction loss over masked modalities, and the contrastive alignment loss:
\[
\mathcal{L} = 
\mathcal{L}_{\text{CE}} 
+ \alpha\,\mathcal{L}_{\text{recon}}
+ \beta\,\mathcal{L}_{\text{con}},
\quad \alpha=\beta=0.1.
\]

In this step, we train for five epochs using the Adam optimizer (learning rate $10^{-3}$, batch size 16); all of the components, including data simulation, encoders, fusion transformer, decoders, contrastive module, and classifier, are implemented in PyTorch to ensure reproducibility.

\begin{table}[t]
\centering
\caption{Key parameters for data generation and model configuration.}
\label{tab:params}
\begin{tabular}{l c}
\toprule
\textbf{Component} & \textbf{Parameter Value} \\
\midrule
EHR sequence length & $T=10$, $d=12$ \\
Imaging resolution & $32\times 32$ \\
Genomics dimension & $500$ features \\
Wearable timeseries & $T=100$, $3$ channels \\
Label noise & $10\%$ flip probability \\
Missing modalities & $30\%$ dropout \\
Embedding dimension & $64$ \\
Fusion transformer & 4-head MHA, 128-unit FFN \\
Classifier & 2-layer MLP + dropout \\
Optimizer & Adam ($10^{-3}$ LR) \\
Batch size & 16 \\
Epochs & 5 \\
\bottomrule
\end{tabular}
\end{table}

\section{Results \& Discussion}

\subsection{Results}

Table~\ref{tab:results} provides a brief summary of the performance of the suggested multimodal foundation model on the held-out test set. The model was trained throughout five epochs, and the optimization was steady as the total loss decreased from 63.42 in the first epoch to 43.56 in the final epoch. Quantitative assessment indicates that the model demonstrates robust predictive performance in the presence of noise and incomplete modalities, as evidenced by balanced classification metrics and elevated area-based scores.

\begin{table}[t]
\centering
\caption{Performance of the multimodal model on the held-out test set.}
\label{tab:results}
\begin{tabular}{l c}
\toprule
\textbf{Metric} & \textbf{Value} \\
\midrule
Training batches & 50 \\
Test batches & 13 \\
Final training loss (Epoch 5) & 43.5574 \\
\midrule
Accuracy & 0.84 \\
Precision & 0.8378 \\
Recall & 0.8692 \\
F1-Score & 0.8532 \\
AUROC & 0.8996 \\
AUPRC & 0.9059 \\
\bottomrule
\end{tabular}
\end{table}

\subsection{Discussion}

The ROC curve (Fig. 1) has a smooth, concave shape and an AUROC of 0.900. This means that the fused representation works well to separate cases of early disease from cases of no disease, even when some modalities are missing or noisy. The precision--recall curve (Fig.~\ref{fig:2}) also showcases high sensitivity, with an AUPRC of 0.906 substantially exceeding the class prior baseline. This suggests that the model maintains reliable precision across a wide range of recall thresholds, which is essential for early detection tasks where false negatives are costly.

\begin{figure}[h]
\centering
\includegraphics[width=0.85\linewidth]{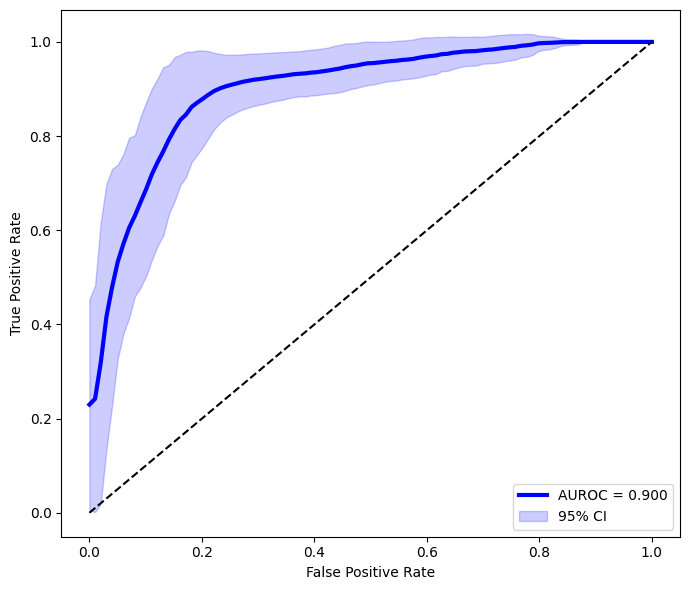}
\caption{ROC curve with 95\% confidence interval (AUROC = 0.900).}
\label{fig:1}
\end{figure}

\begin{figure}[t]
\centering
\includegraphics[width=0.85\linewidth]{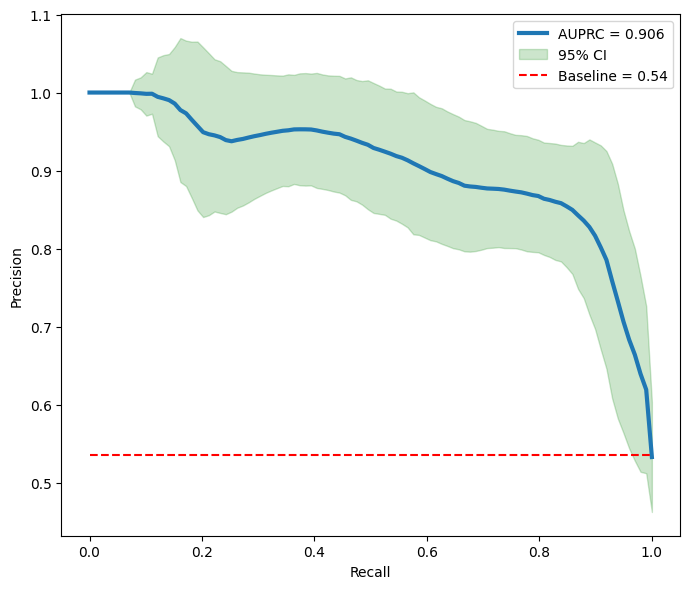}
\caption{Precision--recall curve with 95\% confidence interval (AUPRC = 0.906).}
\label{fig:2}
\end{figure}

Calibration analysis (Fig.~\ref{fig:3})} indicates that predicted probabilities align closely with observed frequencies, with the smoothed trend closely mirroring the ideal diagonal. This means that the uncertainty estimates made by Monte Carlo dropout are very close to the real risk, which means they are good for choosing thresholds and making decisions later on. These trust and transparency issues are similar to blockchain-enabled explainable AI frameworks that make sure clinical AI outputs can be checked and verified \cite{mohsin_blockchainxai_2025}.

\begin{figure}[t]
\centering
\includegraphics[width=0.85\linewidth]{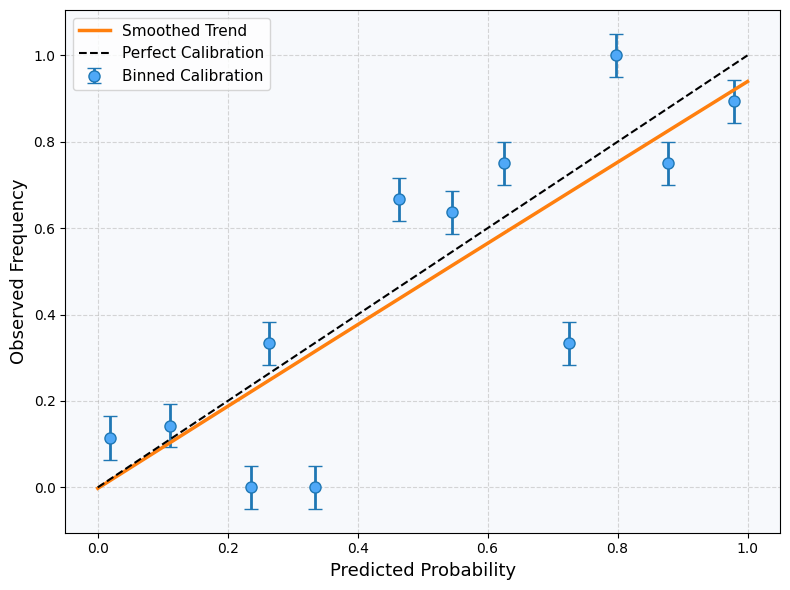}
\caption{Calibration curve showing alignment between predicted probabilities and observed event frequencies.}
\label{fig:3}
\end{figure}

When looked at together, the results show that cross-modal fusion, reconstruction-based self-supervision, and contrastive alignment all make systems more robust to clinical signals that are incomplete or unclear. Although the performance is promising, subsequent efforts will authenticate the framework using actual multimodal patient datasets and investigate the scalability to higher-capacity encoders for more intricate clinical modalities.

\section{Use Cases in Early Disease Detection}

\subsection{Oncology: Multimodal Detection of Cancer}

Doctors still have a hard time finding cancer early because the stage of diagnosis has a direct effect on how long a person lives. Imaging or histopathology are usually used in traditional workflows, but these methods might not be able to find small precancerous or early neoplastic changes on their own. A multimodal foundation model can look at structured electronic health record (EHR) data, pathology slides, genomic variants, and radiological images all at once. Things like your family history and lab results are included. Longitudinal EHR patterns may signify symptom advancement, whereas genomic biomarkers of tumor susceptibility can be assessed in conjunction with radiographic features of alarming lesions. When you put these two signals together, it's easier to find early-stage cancers. There are fewer false negatives, and the screening is more accurate.

\subsection{Cardiovascular Disease: Predicting Heart Failure}

Heart disease is still one of the top causes of death and illness around the world. But it's still very hard to find people who are at high risk early on. Current models often rely on static EHR snapshots, such as echocardiographic results or laboratory values. A multimodal approach can use information from wearable sensors (like heart rate variability and activity levels), cardiac imaging (like echocardiography and MRI), and genomic risk scores to learn more about how a person's body and genes work together. Using these methods together lets you keep checking for risks and encourages quick medical action, which could keep people from going to the hospital and having bad heart outcomes.

\subsection{Neurology: Detection of Neurodegenerative Disorders}

Current models for cardiovascular disease risk prediction often depend on static electronic health record snapshots, which include echocardiographic results or laboratory values. A multimodal approach can use data from wearable sensors (like heart rate variability and activity levels), cardiac imaging (like echocardiography and MRI), and genomic risk scores to give us a better idea of how a person's body works and what their genes are made of. People can keep checking their risk and get medical help right away by using these methods together. This could keep them from going to the hospital and having heart problems.

\section{Conclusion}

We outline a framework for multimodal foundation models designed for healthcare diagnostics. In practice, clinical datasets remain fragmented, frequently include missing modalities, and carry high-dimensional inputs that complicate integration and training {\cite{wu_multimodal_survey_2024}}. Our core idea is to move beyond reliance on a single or fragmented data source, and instead integrate records, images, genomic profiles, and wearable signals into a unified representation. A transformer-based setup is used to handle this mix. Each type of input is first processed by its own encoder, so important details are not lost. These are then linked through attention mechanisms, and pretraining helps the model build general structure before it is tuned for specific clinical tasks. The design also stresses aspects that are often overlooked, like data governance, model monitoring, and interpretability. The framework structure is meant to be flexible across diseases and patient groups. Over time, we hope this kind of system could allow earlier diagnosis, better risk prediction, and treatment tailored to individuals—steps toward precision medicine.

\end{document}